\documentclass[times, twoside]{henriques-style}
\usepackage{blindtext}
\usepackage[nolist]{acronym}
\setcitestyle{square}
\usepackage{tikz}
\usetikzlibrary{positioning, calc, matrix, decorations.pathreplacing, shapes.geometric}
\usepackage{adjustbox}
\usepackage{times}
\usepackage{epsfig}
\usepackage{graphicx}
\usepackage{amsmath}
\usepackage{amssymb}

\usepackage{booktabs}
\usepackage{multirow}

\usepackage{xspace}
\makeatletter
\DeclareRobustCommand\onedot{\futurelet\@let@token\@onedot}
\def\@onedot{\ifx\@let@token.\else.\null\fi\xspace}
\def\etal{\emph{et al}\onedot}

\leadauthor{Jahanifar} 

\begin{document}

\title{Stain-Robust Mitotic Figure Detection for the Mitosis Domain Generalization Challenge}
\shorttitle{TIA Centre MIDOG Algorithm}

\author[1]{Mostafa Jahanifar}
\author[1]{Adam Shephard}
\author[1]{Neda Zamani Tajeddin}
\author[1]{R.M. Saad Bashir}
\author[1]{Mohsin Bilal}
\author[2]{Syed Ali Khurram}
\author[1]{Fayyaz Minhas}
\author[1]{Nasir Rajpoot}

\affil[1]{Tissue Image Analytics Centre, Department of Computer Science, University of Warwick, UK}
\affil[2]{School of Clinical Dentistry, University of Sheffield, Sheffield, UK}

\maketitle


\begin{abstract}
The detection of mitotic figures from different scanners/sites remains an important topic of research, owing to its potential in assisting clinicians with tumour grading. The MItosis DOmain Generalization (MIDOG) challenge aims to test the robustness of detection models on unseen data from multiple scanners for this task. We present a short summary of the approach employed by the \textbf{TIA Centre} team to address this challenge. Our approach is based on a hybrid detection model, where mitotic candidates are segmented on stain normalised images, before being refined by a deep learning classifier. Cross-validation on the training images achieved the F1-score of 0.786 and 0.765 on the preliminary test set, demonstrating the generalizability of our model to unseen data from new scanners.
\end{abstract}



\begin{corrauthor}
mostafa.jahanifar@warwick.ac.uk
\end{corrauthor}


\section*{Introduction}

The detection of mitotic figures is an important task in the analysis of tumour regions \cite{veta2015assessment}. The abundance, or count, of mitotic figures has been shown to be strongly correlated with cell proliferation, which in turn is an important prognostic indicator of tumour behaviour, and thus is a key parameter in several tumour grading systems \cite{veta2015assessment, Aubreville2020}. However, other imposter/mimicker cells are often mistaken for mitotic figures due to their similar appearance/morphology, leading to large inter-rater variability. The introduction of deep learning methods for automated detecting/counting of mitotic figures in histology images offers a potential solution to this challenge.

An additional challenge is the translation of machine learning models into clinical practice (i.e., on whole-slide images or WSIs generated by digital slide scanners), which requires a high degree of robustness to staining and scanner variations. 
The WSIs can vary in their appearance as a result of differences in the way in which the sample was prepared (e.g. preparation/staining procedures) and scanner acquisition method and particular scanner settings. The result of this variation is a {\em domain shift} between WSIs collected from different scanners/sites.

The MItosis DOmain Generalization (MIDOG) challenge \cite{aubreville2021midog} provides a means of testing different algorithms on cohorts of expertly annotated histology images for mitotic figure detection in the presence of a domain shift. To combat these challenges, we first normalise the stain intensities of all images provided to our model before passing images through our proposed hybrid mitosis detection pipeline. The hybrid analysis pipeline consists of (a) a mitotic candidate segmentation model and (b) refinement by a deep learning (DL) classifier. We generated ground truth (GT) segmentation masks of mitotic figures via a semi-automated DL model \cite{jahanifar2019nuclick, koohbanani2020nuclick}.
The use of a pre-trained DL method for GT allows the DL models to exploit the important contextual information and treating this detection task as a segmentation task instead. 



\section*{Methodology}
\label{sec:methods}

\subsection{Image pre-processing}
As stain variation is the dominant challenge when analysing histology images from various scanners, in the first step of our proposed pipeline we used Vahadane \etal's method \cite{vahadane2016structure} to normalise the stain intensities of all images in the training set to the target \textit{image 009}, with the help of TIAtoolbox\footnote{\url{https://github.com/TIA-Lab/tiatoolbox}} library. Note the same stain matrix acquired from \textit{image 009} is used on-the-fly during the prediction on test images. 

\subsection{Mitosis candidate segmentation}
\subsubsection{Mitosis mask generation}
We approach the mitosis candidate detection problem as a segmentation task. However, in order to train a CNN for the segmentation task in a supervised manner, GT masks of the desired objects within the image are required. Since the organizers have only provided approximate bounding box annotations for each mitosis in the released MIDOG dataset, we obtained mitotic instance segmentation masks using NuClick\footnote{\url{https://github.com/navidstuv/NuClick}} \cite{jahanifar2019nuclick, koohbanani2020nuclick}, a CNN-based interactive segmentation model capable of generating precise segmentation masks for each mitotic figure from a point annotation within the mitotic figure. Therefore, for each annotation point in the dataset, we fed the centre point of the bounding box alongside the patch from the original image into NuClick to generate the individual segmentation mask. 


\begin{table*}[!h]
\centering
\begin{tabular}{@{}llllll@{}}
\toprule
                              & $t_{segmentation}$ & $t_{classification}$ & Recall & Precision & F1-Score \\ \midrule
Segmentation only             & 0.5              & -                  & 0.824  & 0.696     & 0.755    \\
Segmentation + Classification & 0.4              & 0.6                & 0.771  & 0.801     & 0.786    \\ \bottomrule
\end{tabular}
\caption{Results of cross-validation experiments on the MIDOG dataset.}
\label{tab:crossval}
\end{table*}

\subsubsection{Segmentation model}
We employed a lightweight segmentation model, called Efficient-UNet \cite{jahanifar2021semantic}, for the segmentation task. The Efficient-UNet is a fully convolutional network based on an encoder-decoder design paradigm where the encoder branch is the B0 variant of Efficient-Net \cite{tan2019efficientnet}. Using this model with pre-trained weights from \textit{ImageNet} as a backbone allows the overall model to benefit from transfer-learning, by extracting better feature representations and gaining higher domain generalizability. The Jaccard loss function \cite{jahanifar2018segmentation} is robust against the imbalanced population of positive and negative pixels in the segmentation dataset, and thus has been utilised to train the model.

\subsubsection{Model training}
In order to train and evaluate the model, we extracted $512\times512$ patches from the stain-normalised images. There was a large class imbalance in the training dataset, owing to the much fewer patches that contained mitosis (positive patches) in comparison to those without mitosis (negative patches). Since we did not wish to introduce a bias towards predicting empty maps (hence increasing the number of false negatives), we devised an on-the-fly under-sampling approach which guaranteed that similar number of positive and negative patches were sampled at the beginning of each epoch. Here, we used all positive patches in all epochs but randomly sampled the negative patches in each epoch. This way we trained a segmentation model that maintains a high level of precision whilst having a high recall.

\subsubsection{Post-processing and candidate extraction}
At the inference stage of the previous step, each image is tiled with overlap ($512\times512$ patches with 75 pixels overlap) and results for all tiles are aggregated to generate the segmentation prediction map.
We then use a sequence of morphological operations and compute the centroid of the connected components to extract candidate mitotic cells from the segmentation map. 

\subsection{Mitosis candidates refinement}
In the final step of our method, the mitosis candidates discovered in the previous step were verified using a classifier. Here, we trained an Efficient-Net-B7 \cite{tan2019efficientnet} classifier to distinguish between mitoses and mimickers. To train the classifier we extracted mitosis and mimicker patches ($96\times96$ pixels) based on the annotations provided by the challenge organizers. Again, to deal with the problem of class imbalance, we incorporated the on-the-fly under-sampling technique. 

\subsection{Data augmentation}
To make both segmentation and classification networks more robust against the variation seen in histology images, we include the standard data augmentation techniques during the network training phase.
The extent and combinations of these augmentation techniques are randomly selected on-the-fly and differ from epoch to epoch.

\subsection{Inference}
\label{inference}
The same pipeline as used for training was applied to each input image for inference. However, in order to benefit from all the models and all the training data, 
we also included ``model ensembling'' and ``test time augmentation'' (TTA) techniques in the inference pipeline. Therefore, during segmentation and classification, predictions from all three models from the cross-validation experiments (in addition to predictions on input image variations by TTA techniques like image flipping and sharpening) are averaged to make more confident and robust final predictions on unseen data.

\section*{Evaluation and Results}

The training set released with the MIDOG challenge contains 150 images with GT annotations. 
All segmentation and classification models were evaluated in a cross-validation framework as follows: three folds were created based on the images from different scanners (fold 1: images 1-50; fold 2: images 51-100; fold 3: images 101-150). Three experiments were conducted, where the models were trained on two folds and validated on the final fold, such that all images were tested once. Our training scheme simulated the way in which the challenge is tested i.e. the test scanner is not used during the network training.

Many configurations for the segmentation and classification networks were tested, but the ones with minimum segmentation loss and best classification F1-score on the validation set were selected. In \cref{tab:crossval}, results of the cross-validation experiments for the segmentation only and the hybrid (segmentation+classification) models are reported separately. The segmentation model alone achieved a F1-score of 0.755 in mitosis detection over all the images in the training set. In comparison, the addition of the classier (mitosis candidate refinement step) increased the F1-score to 0.786. Note that the threshold values and hyper parameters on each step in the proposed pipeline are selected based on the cross-validation experiments.

Using the proposed hybrid method and the above inference method, we were able to achieve the F1-score of 0.765 on the preliminary test set. This small reduction in F1-score on testing, when compared to cross-validation, indicates the robustness and generalizability of the proposed approach.

\section*{Discussion and Conclusion}

In this work, we have presented a new method for the challenge of mitotic figure detection in histology images in the presence of a domain shift. Our proposed method first segments mitotic figures, based on the Efficient-UNet architecture, before passing the results of segmentation on to a DL-based classifier to further differentiate between mitotic figures and hard negatives (mimickers). All images were normalised to a chosen sample image during training, before being normalised on-the-fly during inference. The proposed method achieved a high F1-score of 0.765 when tested on the preliminary test set for the MIDOG challenge.

\section*{Bibliography}

{\small
\bibliography{egbib}
}

\end{document}